\documentclass[review]{elsarticle}

\usepackage{lineno,hyperref}
\usepackage{graphicx}
\usepackage{ dsfont }
\usepackage{graphicx}
\usepackage{amsmath}
\usepackage{xcolor}
\usepackage[misc,geometry]{ifsym} 
\usepackage{amsmath,soul}
\usepackage{amsfonts}
\usepackage[english]{babel}
\usepackage[utf8]{inputenc}
\usepackage{makecell}
\usepackage{hyperref}
\newcommand{\bftab}{\fontseries{b}\selectfont}
\modulolinenumbers[5]
\usepackage{rotating}
\usepackage{color, colortbl}
\usepackage{algorithm,algpseudocode}
\usepackage{caption}
\usepackage{listings}
\usepackage{ulem}

\journal{Neural Networks}

\definecolor{light-gray}{gray}{0.90}  
\makeatletter
\def\ps@pprintTitle{%
   \let\@oddhead\@empty
   \let\@evenhead\@empty
   \def\@oddfoot{\reset@font\hfil\thepage\hfil}
   \let\@evenfoot\@oddfoot
}
\makeatother

\begin{document}

\begin{frontmatter}

\title{Weakly Supervised Thoracic Disease Localization via Disease Masks}






\author[address_1]{Hyun-Woo Kim\fnref{equal}}
\author[address_1]{Hong-Gyu Jung\fnref{equal}}
\author[address_1,address_2]{Seong-Whan Lee\corref{mycorrespondingauthor}}
\fntext[equal]{Equal contribution}
\cortext[mycorrespondingauthor]{Corresponding author}
\ead{sw.lee@korea.ac.kr}

\address[address_1]{Department of Brain and Cognitive Engineering, Korea University, Anam-dong, Seongbuk-gu, Seoul, 02841, Korea}
\address[address_2]{Department of Artificial Intelligence, Korea University, Anam-dong, Seongbuk-gu, Seoul, 02841, Korea}

\begin{abstract}
To enable a deep learning-based system to be used in the medical domain as a computer-aided diagnosis system, it is essential to not only classify diseases but also present the locations of the diseases. However, collecting instance-level annotations for various thoracic diseases is expensive. Therefore, weakly supervised localization methods have been proposed that use only image-level annotation. While the previous methods presented the disease location as the most discriminative part for classification, this causes a deep network to localize wrong areas for indistinguishable X-ray images. To solve this issue, we propose a spatial attention method using disease masks that describe the areas where diseases mainly occur. We then apply the spatial attention to find the precise disease area by highlighting the highest probability of disease occurrence. Meanwhile, the various sizes, rotations and noise in chest X-ray images make generating the disease masks challenging. To reduce the variation among images, we employ an alignment module to transform an input X-ray image into a generalized image. Through extensive experiments on the NIH-Chest X-ray dataset with eight kinds of diseases, we show that the proposed method results in superior localization performances compared to state-of-the-art methods.

\end{abstract}
\end{frontmatter}

\section{Introduction}

Obtaining a precise diagnosis is critical for early treatments of many diseases and ensuring successful recovery. Thus, radiologists must analyze large numbers of X-ray images and make diagnosis quickly with high precision. To reduce these efforts, computer-aided diagnosis (CAD) tools for X-ray tests have become an essential element for providing a second opinion to radiologists. Radiologists can use the CAD tools to reduce the number of false positive cases and achieve accurate diagnoses in less time. Along the same line, deep learning has been successfully used in a variety of research fields~\cite{he2017mask,anderson2018bottom,ren2015faster,he2016deep} to replace handcrafted feature approaches~\cite{kang2014nighttime, roh2007accurate, roh2010view} and many studies have investigated applying deep learning to the medical domain to help radiologists diagnose disease~\cite{lanfredi2019adversarial,tang2019tuna,xue2018multimodal,liu2019clinically,wang2018tienet,taghanaki2019infomask,xue2019improved,zhang2020radiology, peng2020discretely, kim2019multi}. As a result, there has been considerable progress in thoracic disease classification. 

On the other hand, it is difficult for radiologists to trust the prediction results when an accurate location related to the diagnosed disease is not provided. Thus, localizing disease areas is a prerequisite for using deep learning methods as tools for providing a second opinion. However, the machine approaches are still inadequate at providing the location of thoracic disease due to the lack of instance-level annotation data in the medical field. Specifically, not only radiologists must laboriously analyze X-ray images for extended periods to generate the instance-level information as training data, but also deep learning requires considerable training data of the disease areas. 

\begin{figure}[t]
\centering
\includegraphics[width=1.0\columnwidth]{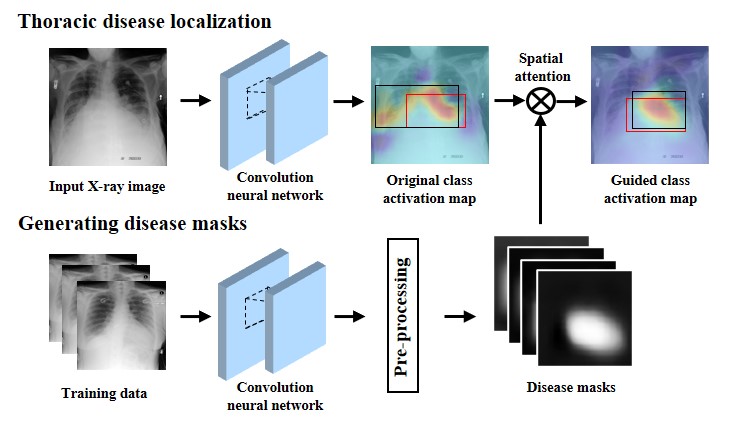}
\vspace*{-5mm}
\caption{Overview of the proposed spatial attention with disease masks. The original class activation map generated from a deep network localizes the disease in the wrong area. However, disease masks can guide the class activation map to follow the distribution of the disease area. The red and black boxes indicate the ground truth and predicted areas, respectively.}\label{fig1}
\end{figure}

To solve this problem, we propose a weakly supervised disease localization method that uses only image-level information to reduce the high cost incurred by detailed annotations. Because hospitals already have access to large numbers of X-ray images for thoracic diseases, a weak supervision approach simplifies the problem and is easier than collecting instance-level annotations individually. Various studies have been conducted to investigate weakly supervised thoracic disease localization~\cite{wang2017chestx, cai2018iterative, li2018thoracic, liu2019align}. These previous works have traditionally presented disease areas using activated feature maps from a classifier. However, they do not consider the typical disease occurrence areas, and thus they often fail to localize the disease area as recognized by radiologists. Each thoracic disease has an area where it most frequently occurs. For example, cardiomegaly occurs around the heart, while pneumonia occurs in the lungs. Intuitively, radiologists do not look at all the areas when diagnosing thoracic X-ray images. Instead, they concentrate on the part related to the patient's symptoms. However, the existing studies visualize only the most discriminative areas for the disease classification of a single subject without considering prior knowledge regarding the area where a disease most frequently occurs. 

To address this issue, we propose a spatial attention method using disease masks containing the spatial probability of disease occurrence. Fig. \ref{fig1} shows an overview of the spatial attention with disease masks for thoracic disease localization. The disease masks are generated by a pretrained network using X-ray training data and the activated features. The masks are used to guide the network to detect the disease area more accurately.

Meanwhile, X-ray data have problems due to their diverse variations, such as rotation and shift. Specifically, the diverse variations in X-ray images can prevent the disease mask from focusing the spatial attention on the common areas where diseases occur. For this reason, we use an alignment module \cite{liu2019align} to transform the X-ray image into a generalized chest shape. In addition, medical datasets typically have large imbalances in the number of positive and negative samples\footnote{Positive and negative samples indicate whether a X-ray image contains diseases or not.}, causing classifiers to have a skewed predictions. To alleviate this problem, we train the network using a weighted loss function \cite{wang2017chestx} that assigns a larger weighted loss to the classes with fewer training examples. 

In the experimental section, we compare the proposed method with previous works~\cite{wang2017chestx, cai2018iterative, liu2019align} on the NIH-Chest X-ray dataset. The results show that our method achieves state-of-the-art localization performance. In addition, we demonstrate the effectiveness of the generated disease masks through extensive ablation studies. In summary, our contributions are as follows:

\begin{itemize}
\item We present a spatial attention method using disease masks that provide prior knowledge about disease occurrence areas.
\item We propose a unified framework using input alignment, feature attention and a loss function that compensates for imbalanced chest X-ray datasets.
\item We show that our proposed method achieves state-of-the-art performance for weakly-supervised thoracic disease localization.
\end{itemize}

\section{Related Works}
\subsection{Weakly supervised localization}\label{sect:wsl}

Recently, weakly supervised localization methods using class activation map (CAM)~\cite{zhou2016learning} and Grad-CAM~\cite{selvaraju2017grad} approaches that help explain the reasons for the class judgments of classifiers have received considerable attention. In~\cite{zhou2016learning}, B. Zhou \textit{et al}. proposed CAM to visualize which features a network uses to make decisions by using the weights and feature maps from the final layer. In~\cite{selvaraju2017grad}, R.R. Selvaraju \textit{et al}. proposed Grad-CAM, which visualizes the crucial features using gradient information obtained during the back-propagation process. Unlike CAM, Grad-CAM is applicable to multiple types of networks such as those used for image captioning~\cite{chen2015microsoft, fang2015captions} and visual question answering~\cite{antol2015vqa, gao2015you} as well as for networks applied to image classification tasks. However, CAM and Grad-CAM both visualize only the most discriminative area rather than the complete object areas used to classify the data. 

To overcome this limitation, various weakly supervised object localization methods have been proposed~\cite{kumar2017hide,zhang2018adversarial,zhang2018self,choe2019attention,zhou2019dual} that detect an object's location using only image-level information. In~\cite{kumar2017hide}, K. Singh \textit{et al}. proposed a grid drop method in which an input image is divided into an $n$-by-$n$ grid, and each grid cell is randomly dropped with a fixed probability. Then, the modified image is repeatedly used for training, enabling a network to find the entire object area. However, the random grid drop method requires a long training time. In~\cite{zhang2018adversarial}, X. Zhang \textit{et al}. proposed a method to localize object areas using two classifiers that share a single feature extractor. The first classifier finds the most discriminative area and removes the features from that area. Then, the second classifier trains the network using the removed features and finds the next most discriminative areas. Finally, they combined the class activation maps generated by the two classifiers to perform object localization.  In~\cite{zhang2018self}, X. Zhang \textit{et al}. proposed a self-produced guidance (SPG) mask that divides an object into foreground and background areas. They trained a network with the SPG mask to indicate the spatial correlation of locations. In~\cite{choe2019attention}, J. Choe \textit{et al}. proposed a method to train a network by randomly dropping the most strongly activated region in a self-attention mask. 

Although the algorithms are mainly developed using attention-based methods, they does not consider the characteristic of medical images where  there exist indistinguishable X-ray images with different diseases or a small disease.

\subsection{Thoracic disease localization}
As a baseline, X. Wang \textit{et al}. published ChestX-ray14~\cite{wang2017chestx} that is a large hospital-scale dataset and they studied the classification and localization problems of thoracic diseases in a multilabel environment. In~\cite{cai2018iterative}, J. Cai \textit{et al}. proposed a two-step approach. Using an attention-mining (AM), they generated first activation maps, and second activation maps were produced after removing the activated area from the first one. Then, they used the L2 distance to consider the relationships between multiple diseases within an image. In~\cite{li2018thoracic}, L. Zhe \textit{et al}. addressed that it is challenging to collect instance-level annotations for all subjects to localize thoracic disease. They proposed a network structure and a conditional loss function that can be trained simultaneously from small amounts of annotated and unannotated data. In~\cite{liu2019align}, J. Liu \textit{et al}. pointed out that the ChestX-ray14 dataset lacks high-quality images. They proposed an alignment module and a contrast-induced attention network that increased the attention to the disease area by contrasting negative images with similar positive images. 

However, the previous approaches rely on the discriminative power of deep networks that are trained end-to-end for classification without considering the disease occurrence area. Instead, we propose an attention method using disease masks to guide the network to consider the typical spatial distributions of thoracic diseases.

\section{Methods}

\begin{figure}[t]
\centering
\includegraphics[width=1 \columnwidth]{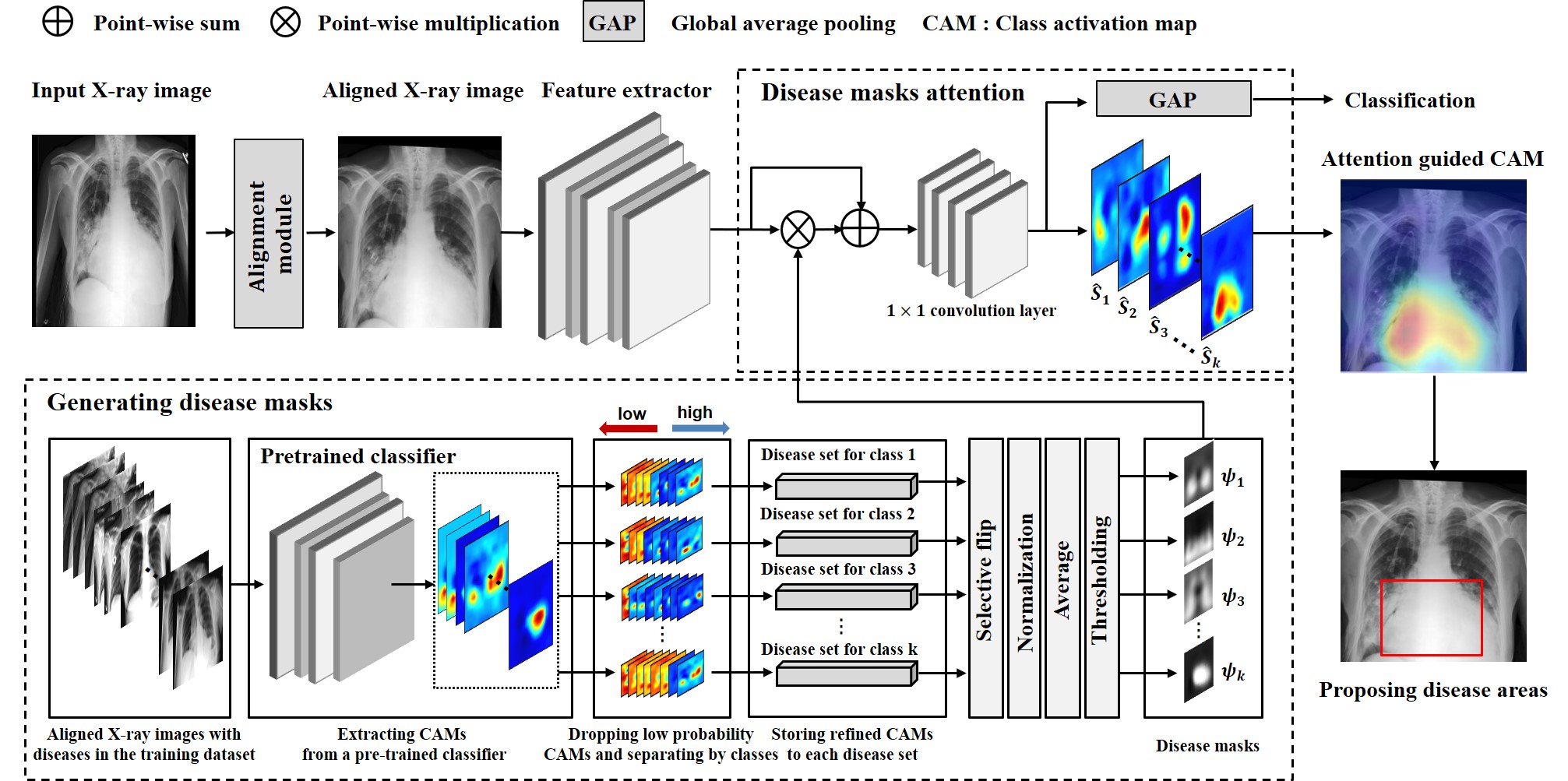}
\vspace{-5mm}
\caption{Detailed overview of the DM attention architecture. The DM attention method consists of two parts: generating disease masks and focusing the spatial attention of deep networks using the masks. The disease masks are generated using aligned X-ray images and a pretrained classifier, and each map indicates the representative area for a disease.}\label{fig2}   
\end{figure}

\subsection{Spatial attention with disease masks}\label{sect:main}
As shown in Fig.~\ref{fig2}, the proposed method consists of two parts by generating disease masks (DM) and providing spatial attention to the localization network using the masks with a high probability of disease occurrence. We utilize the class activation maps using a pretrained network to generate the disease masks representing each class and retrain the feature extractor by fusing the generated disease masks and extracted feature maps. As a result, the retrained feature extractor can localize diseases by considering the occurrence distribution for each disease. The details of each task are described in following subsections.


\subsubsection{Generating disease masks}\label{sect:dm}
To generate disease masks, we extract class activation maps~\cite{zhou2016learning} using aligned X-ray images from a pretrained network. Let $x \in \mathbb{R}^{P\times P \times 3}$ detnote aligned X-ray images where $P$ defines the dimensions of the image. The $c$-th class activation map $S_{c}(x) \in \mathbb{R}^{p\times p\times 1}$ is defined as follows:
\begin{equation}\label{eq0}
    S_{c}(x) = \sum_{k=1}^{C'}w_{k}^{c}f_{k}\left(x\right),
\end{equation}
where $f (x) \in \mathbb{R}^{p\times p\times C'}$ denotes the extracted $C'$ feature maps of size $p$ from a pretrained network and $w_{k}^{c}$ is the weight of the $k$-th feature map for class $c$. Then, the probability score of $x$ in the $c$-th class is defined as
\begin{equation}\label{eq0_1}
    P_{c}(x) = Sigmoid\left(\sum_{j=1}^{p}\sum_{i=1}^{p}S_{c}^{i,j}\left(x\right)\right),
\end{equation}
where $P_{c}(x)$ is obtained using the sigmoid function after performing global averaging pooling on $S_{c}(x)$. Then, we collect $\bar{x}$ by dropping images $x$ where $P_{c}(x)<0.8$ to obtain high-quality class activation maps. 

Finally, we generate the class-specific disease masks using the activation maps. The disease mask for class $c$ is defined as follows: 
\begin{equation}\label{eq1}
	\psi_{c}=T\left({1 \over N_{c}}\sum_{d=1}^{N_{c}}N\left(\bar{L}^{d}_{c}\right)\right) \mbox{where}\; \bar{L}^{d}_{c}  = \begin{cases}
                & L_{c}^{d},\ \mbox{if}\ c = m\\
                & H\left( L_{c}^{d}\right)+ L_{c}^{d},\ \mbox{otherwise.}
                \end{cases}
\end{equation}
Here, $L_c^d$ represent the $d$-th element of $L_{c}=\{S_{c}(\bar{x}_{1}),S_{c}(\bar{x}_{2}), \cdots,S_{c}(\bar{x}_{N_{c}})\}$, and $N_c$ is the number of high-quality images $\bar{x}$ for class $c$. In addition, $m$ denotes an asymmetrically located class, and $H(\cdot)$ is the horizontal flip function. We normalize the activation maps to bring all values into the range $[0,1]$ using $N(\cdot)$ and apply a thresholding function $T(\cdot)$ to obtain the highly activated areas.

\begin{figure}[t!]
\centering
  \includegraphics[width=0.8\columnwidth]{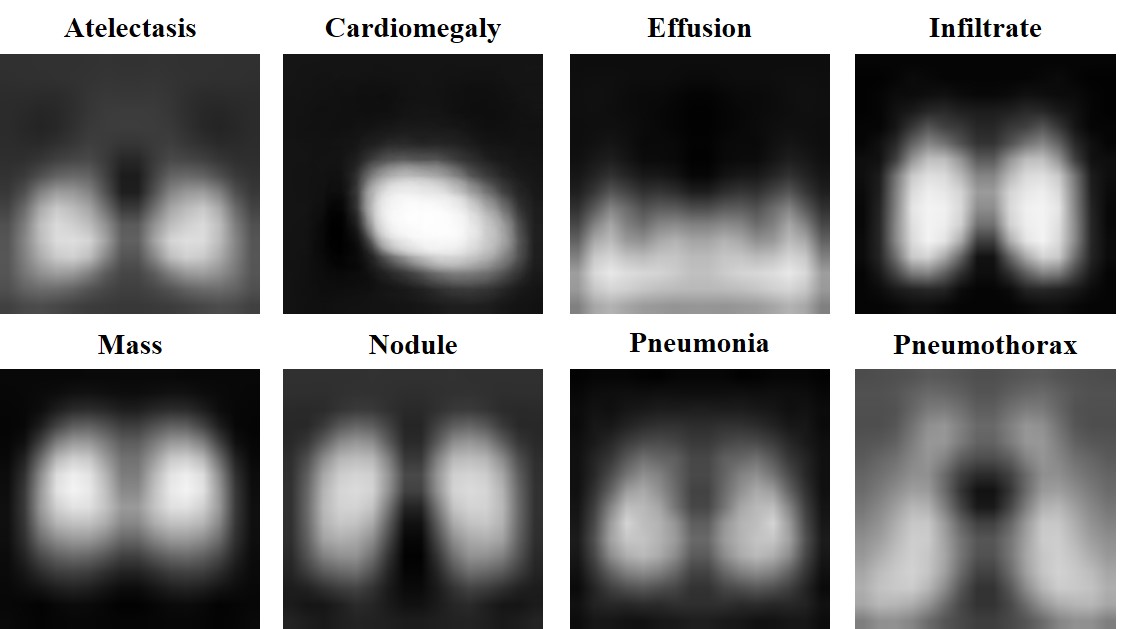}
   \caption{The generated disease masks on the eight kinds of diseases. Each mask describes an area where diseases mainly occur. The disease masks are used to give spatial attention to a deep network.
}\label{fig6}
\end{figure}

Specifically, seven types of diseases (atelectasis, effusion, infiltrate, mass, nodule, pneumonia and pneumothorax) mainly occur symmetrically in the lungs. On the other hand, cardiomegaly is a disease in which the left ventricle area is enlarged and occurs asymmetrically. To reflect the innate characteristics of these diseases, we apply $H(L_{c}^{d})+L_{c}^{d}$ for the seven symmetrical diseases and $L_{c}^{d}$ for the asymmetrical disease. Fig.~\ref{fig6} shows the generated disease masks for the eight kinds of diseases considered in this study. 

\subsubsection{Localization with spatial attention}\label{sect:sa}
We retrain the feature extractor with the disease masks so that the deep network focuses on the areas where diseases frequently occur. To employ spatial attention, attention guided feature maps are generated as follows:
\begin{equation}\label{eq2_0}
f'\left( x \right) = \psi \cdot f\left( x \right) + f\left( x \right).
\end{equation}
The disease masks deactivate the parts with a lower probability of disease occurrences. Moreover, the skip connections effectively combine the information from the attention and the original feature maps. 

After retraining, we generate class activation maps using the attention guided class activation map $ \hat{S}_c(x) \in \mathbb{R}^{p \times p \times 1}$ as follows:
\begin{equation}\label{eq2}
\hat{S}_c(x)= \sum_{k=1}^{C'}w_{k}^{c}{f'_{k}\left(x\right)}.
\end{equation}

Finally, we employ the class activation map to localize the thoracic disease areas. The essence of spatial attention is to reinforce weakly activated areas by guiding the deep network to refer to the innate properties of thoracic diseases.

\begin{figure}[t!]
\centering
\includegraphics[width=1.0\columnwidth]{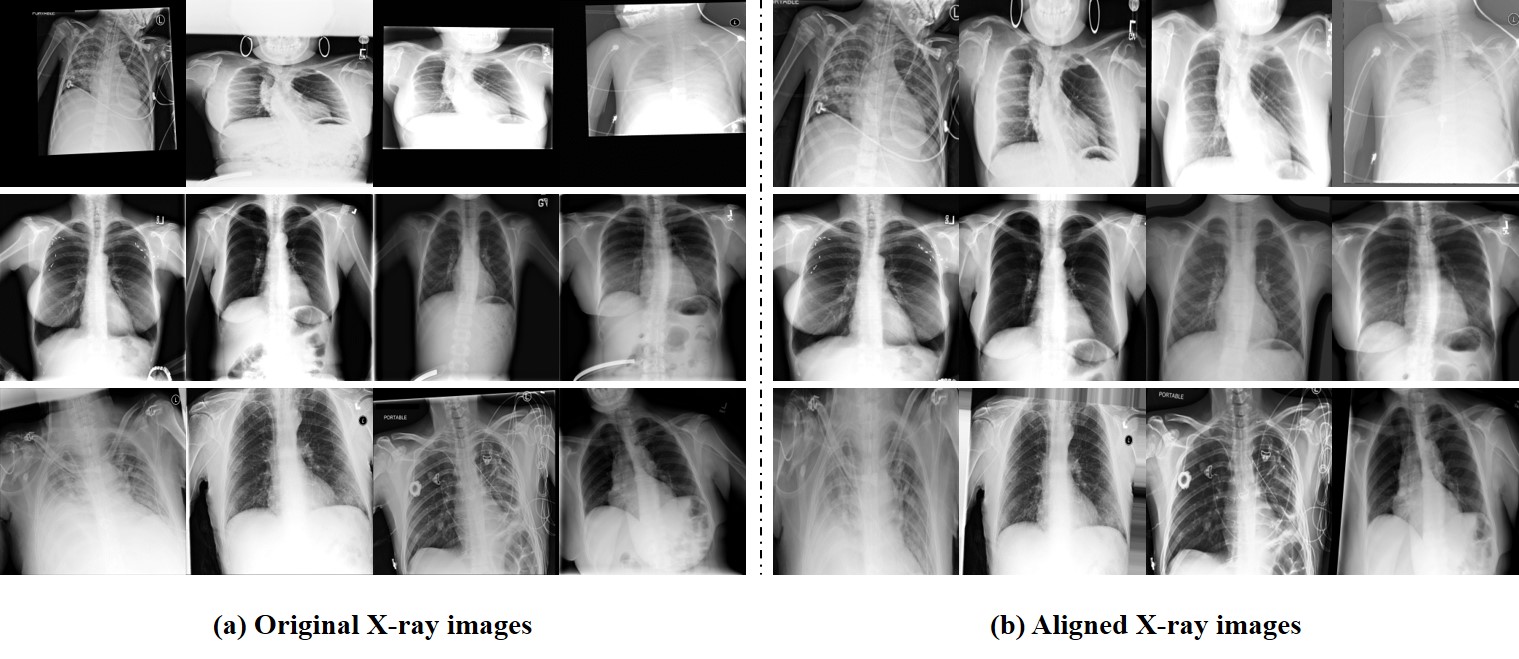}
\vspace{-8mm}
\caption{Examples of original X-ray images and aligned X-ray images. (a) The incorrectly aligned original X-ray images. (b) The improved X-ray images from the alignment module.
}\label{fig5}
\end{figure}

\subsection{Alignment module} \label{sect:AM}
The original X-ray image generally contains diverse types of variations such as rotations, shifts, and different scales. These misaligned X-ray images are commonly found in various chest X-ray datasets \cite{wang2017chestx, irvin2019chexpert, johnson2019mimic} and make it hard to find the disease occurrence areas for generating the disease masks. Moreover, these variations function as noise in the deep network, making it challenging to identify disease and causing the proposed disease masks to guide the network to the wrong area. 

To solve this problem, we use an alignment module \cite{liu2019align} that reduces the variations in the original X-ray images. The alignment module applies an affine transformation to the original X-ray images to generate aligned X-ray images. Fig.~\ref{fig5} shows some examples of original and aligned X-ray images. In Fig.~\ref{fig5}(a), the original X-ray images contain diverse variations, but Fig.~\ref{fig5}(b) shows higher-quality X-ray images obtained from the alignment module with effectively reduced variations.
\begin{figure*}[t!]
\centering
\includegraphics[width=1\columnwidth]{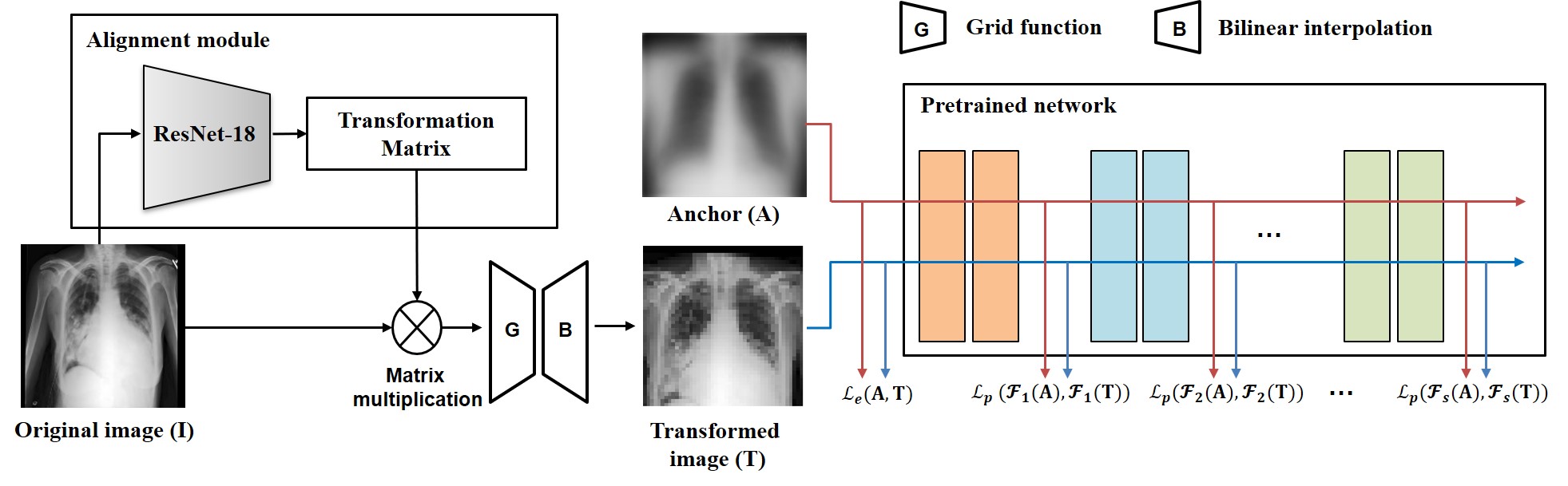}
\vspace{-6mm}
\caption{The optimization process of the alignment module with perceptual loss and Euclidean loss. The alignment module is trained to transform the original X-ray image similarly to the anchor image.}\label{fig3}
\end{figure*}

Fig.~\ref{fig3} shows the optimization process of the alignment module. The anchor image $A \in \mathbb{R}^{P \times P \times 3}$ is a generalized chest image generated from the average of 2,000 normal random samples. The alignment module is implemented using ResNet-18~\cite{he2016deep}, and it outputs five parameters ($s_{x}$, $s_{y}$, $t_{x}$, $t_{y}$, $\theta$) where $s_{x}$ and $s_{y}$ denote the horizontal and vertical scales, respectively, $t_{x}$ and $t_{y}$ denote the horizontal and vertical displacements, respectively, and $\theta$ denotes the rotation angle. The transformed image $\varphi$($I$) is defined as follows:
\begin{equation}\label{eq3}
    \varphi(I)=B \left(G \left(
    \left[
    \begin{matrix}s_{x}
    cos\theta & -s_{y} sin\theta & t_{x}\\
    s_{x}sin\theta & s_{y}cos\theta&t_{y}\\ 
    \end{matrix} 
    \right]
    I\right) \right),
\end{equation}
where $I \in \mathbb{R}^{P \times P \times 3}$ indicates an input X-ray image and the five parameters are the outputs from the alignment module. We reduce the size of the transformed image using a grid function $G\left( \cdot \right)$ that performs the average pooling with a kernel size of 16 and stride of 16. Then, we restore the size using billinear interpolation $B\left( \cdot \right)$. It is worth noting that averaging the 2,000 samples reduces the noise of various shapes but causes the anchor image to be blurred. Thus, the $B\left(G\left(\cdot\right)\right)$ process smooths the transformed image so that it more closely  resembles the blurred anchor image.

We train the alignment module to make $\varphi(I)$ as similar as possible to the shape of the anchor image using perceptual loss~\cite{johnson2016perceptual} and Euclidean loss as in \cite{liu2019align}. The perceptual loss is defined as follows:
\begin{equation}\label{eq4}
\mathcal{L}_{p}\left(A, I\right)=\sum_{s=1}^{N_{I}}\frac{1}{C^{s}H^{s}W^{s}}||\mathcal{F}_{s}\left(A\right)-\mathcal{F}_{s}\left(\varphi\left(I\right)\right)||_2,
\end{equation}
where $\mathcal{F}_{s} \left( \cdot \right)$ represents the intermediate features of the $s$-th layer extracted from the pretrained network and $C^{s}, H^{s}, W^{s}$ represents the feature size. The perceptual loss function penalizes the difference between the intermediate features of the two images $A$ and $\varphi(I)$. In addition, the Euclidean loss that penalizes the difference between two images $A$, $\varphi(I)$ is defined as
\begin{equation}\label{eq5}
\mathcal{L}_{e}\left(A,I\right)=\sum_{j=1}^{P}\sum_{i=1}^{P}||A^{i,j}-\varphi\left(I\right)^{i,j}||_{2}.
\end{equation}

Finally, the total loss used to train the alignment module is as follows:
\begin{equation}\label{eq6}
\mathcal{L}_{total}=L_{e}+L_{p}.
\end{equation}
Thus, we aim to learn the transformation parameters from a pretrained network by considering both the images and their complex features.

\subsection{Weighted cross-entropy loss}
Medical datasets typically have a data imbalance problem between the positive and negative samples. This imbalanced data causes a classifier to output predictions skewed to specific class. To solve this problem, we adopt weighted binary cross-entropy~\cite{wang2017chestx}, which is defined as follows:
\begin{equation}\label{eq7}
	\mathcal{L}_{WCE}=\omega_{P}\sum_{y \in P}-ln\left(M\left(x\right)\right)+\omega_{N}\sum_{y \in N}-ln{\left(1-M\left(x\right)\right)},
\end{equation}
where $x$ and $y$ represent the aligned X-ray images and the corresponding labels, respectively, $M\left( \cdot \right)$ indicates the prediction scores of a classifier, $\omega_{P}$ and $\omega_{N}$ are denoted as $\omega_{P}=\frac{|P|+|N|}{|P|}$, $\omega_{N}=\frac{|P|+|N|}{|N|}$. $|P|$, and $|N|$ represents the number of positive and negative samples in a batch.  When many positive samples exist in the batch, the value of $\omega_{P}$ decreases proportionatrely, while the value of $\omega_{N}$ increases. In contrast, when many negative samples exist in the batch, the value of $\omega_{N}$ decreases proportionately, and the value of  $\omega_{P}$ increases. 


\section{Experiments}
\subsection{Dataset and preprocessing}
A previous study~\cite{wang2017chestx} published the ChestX-ray14 dataset which can be used to identify and localize thoracic diseases. This dataset consists of 112,120 frontal-view chest X-rays and contains the image-level annotation as 86,524 for training and validation and 25,586 for testing, and it contains image-level annotations in which each image is multilabeled for 14 types of diseases (atelectasis, cardiomegaly, effusion, infiltrate, mass, nodule, pneumonia, pneumothorax, consolidation, edema, emphysema, fibrosis, pleural thickening and hernia). Moreover, the dataset provides 984 bounding box annotations for eight kinds of diseases (atelectasis, cardiomegaly, effusion, infiltrate, mass, nodule, pneumonia and pneumothorax). We resized the original X-ray images from 1024$\times$1024$\times$3  to 512$\times$512$\times$3 to enable faster training and normalized each channel to [-1,1].

\subsection{Hyperparameter setting}
Our architecture consists of the alignment module and the attention network. These two networks were trained separately. The alignment module was trained for five epochs using ResNet-18 to generate an affine transform matrix. To extract intermediate features in the perceptual loss, we exploited the `conv2\_x', `conv3\_x', `conv4\_x' and `conv5\_x' layers from a ResNet-50 model pretrained on the ChestX-ray14 dataset. We used the stochastic gradient descent algorithm with a learning rate of 0.001 and a batch size of 16. We then trained our feature extractor with ResNet-50 using the aligned X-ray images for five epochs and then retrained it for five more epochs with generated disease masks. We used the Adam optimizer~\cite{kingma2014adam} with a learning rate and batch size of 0.0001 and 32, respectively. 

\subsection{Evaluation of thoracic disease localization}
We used the accuracy metric to evaluate localization performances. A higher accuracy score implies a better localization model. This metric was evaluated using the Intersection over Union (IoU) of the overlapping region between the predicted boxes and the ground truth. The accuracy metric predicts true positives when the IoU of two boxes is above a threshold. We evaluated thoracic disease localization accuracy by setting thresholds of [0.3, 0.5, 0.7]. To change the class activation map into a measurable prediction box, we perform binarization of activation maps using the predefined threshold presented in~\cite{wang2017chestx, ye2020weakly}.

\begin{table}[t]
\caption{Comparison of localization performance according to IoU thresholds. Our proposed model (ResNet-50+Alignment+DM) shows state-of-the-art localization performance for eight kinds of thoracic diseases. In addition, the DM attention method shows improved localization performance compared with the ResNet-50 and ResNet-50+Alignment models.
\label{tab1}}
\centering
\resizebox{\textwidth}{!}{
\begin{tabular}{l|l|llllllll|l}
\Xhline{2\arrayrulewidth}
\multicolumn{1}{c|}{T(IoU)} & \multicolumn{1}{c|}{Model} & \multicolumn{1}{c}{Ate.} & \multicolumn{1}{c}{Car.} & \multicolumn{1}{c}{Eff.} & \multicolumn{1}{c}{Inf.} & \multicolumn{1}{c}{Mas.} & \multicolumn{1}{c}{Nod.} & \multicolumn{1}{c}{Pn1.} & \multicolumn{1}{c|}{Pn2.} & \multicolumn{1}{c}{Mean} \\ \Xhline{2\arrayrulewidth}
\multicolumn{1}{c|}{0.3}    & X. Wang \textit{et al}.~\cite{wang2017chestx} & 0.24 & 0.46 & 0.30 & 0.28 & 0.15 & 0.04 & 0.17 & 0.13 & 0.22 \\ 
                             & J. Cai \textit{et al}.~\cite{cai2018iterative}      & 0.33 & 0.85 & 0.34 & 0.28 & 0.33 & 0.11 & 0.39 & 0.16 & 0.35 \\ 
                             & J. Liu \textit{et al}.~\cite{liu2019align}  & 0.34 & 0.71 & 0.39 & \bftab{0.65} & 0.48 & 0.09 & 0.16 & 0.20 & 0.38 \\ \cline{2-11}
                             & ResNet-50                 & 0.30 & 0.82 & 0.32 & 0.21 & 0.35 & 0.10 & 0.19 & 0.20 & 0.31 \\ 
                             & ResNet-50+Alignment      & 0.37 & 0.80 & 0.37 & 0.41 & 0.40 & 0.11 & 0.25 & 0.22 & 0.37 \\ 
                             & \cellcolor{light-gray}ResNet-50+Alignment+DM    &\cellcolor{light-gray}\bftab{0.40} & 
                                                            \cellcolor{light-gray}\bftab{0.97} &
                                                            \cellcolor{light-gray}\bftab{0.40} &
                                                            \cellcolor{light-gray}0.58 &
                                                            \cellcolor{light-gray}\bftab{0.50} &
                                                            \cellcolor{light-gray}\bftab{0.14} & 
                                                            \cellcolor{light-gray}\bftab{0.40} & 
                                                            \cellcolor{light-gray}\bftab{0.38} & 
                                                            \cellcolor{light-gray}\bftab{0.47}            
                            \\ \hline \hline
\multicolumn{1}{c|}{0.5}    & X. Wang \textit{et al}.~\cite{wang2017chestx} & 0.05 & 0.18 & 0.11 & 0.07 & 0.01 & 0.01 & 0.03 & 0.03 & 0.06 \\
                             &J. Cai \textit{et al}.~\cite{cai2018iterative} & 0.11 & 0.60 & 0.10 & 0.12 & 0.07 & 0.03 & \bftab{0.17} & 0.08 & 0.16 \\
                             & J. Liu \textit{et al}.~\cite{liu2019align}   & \bftab{0.19} & 0.53 & 0.19 & \bftab{0.47} & 0.33 & 0.03 & 0.08 & 0.11 & 0.24 \\ \cline{2-11}
                             & ResNet-50                  & 0.10 & 0.60 & 0.13 & 0.15 & 0.13 & 0.01 & 0.10 & 0.12 & 0.17 \\
                             & ResNet-50+Alignment       & 0.17 & 0.64 & 0.20 & 0.18 & 0.28 & 0.03 & 0.09 & 0.11 & 0.21 \\
                             & \cellcolor{light-gray}ResNet-50+Alignment+DM    &\cellcolor{light-gray}0.18 &
                                                            \cellcolor{light-gray}\bftab{0.80} &
                                                            \cellcolor{light-gray}\bftab{0.21} &
                                                            \cellcolor{light-gray}0.25 &
                                                            \cellcolor{light-gray}\bftab{0.35} & 
                                                            \cellcolor{light-gray}\bftab{0.05} & 
                                                            \cellcolor{light-gray}0.15& 
                                                            \cellcolor{light-gray}\bftab{0.16} &
                                                            \cellcolor{light-gray}\bftab{0.27}
                            \\ \hline \hline
\multicolumn{1}{c|}{0.7}    & X. Wang \textit{et al}.~\cite{wang2017chestx} & 0.01 & 0.03 & 0.02 & 0.00 & 0.00 & 0.00 & 0.01 & 0.02 & 0.01 \\
                             & J. Cai \textit{et al}.~\cite{cai2018iterative} & 0.01 & 0.17 & 0.01 & 0.02 & 0.01 & 0.00 & 0.02 & 0.02 & 0.03 \\ 
                             & J. Liu \textit{et al}.~\cite{liu2019align}  & \bftab{0.08} & 0.30 & 0.09 & \bftab{0.25} & \bftab{0.19} & \bftab{0.01} & 0.04 & 0.07 & \bftab{0.13} \\ \cline{2-11}
                              & ResNet-50                 & 0.06 & 0.21 & 0.04 & 0.07 & 0.03 & 0.00 & 0.02 & 0.03 & 0.06 \\
                             & ResNet-50+Alignment      & \bftab{0.08} & 0.35 & 0.10 & 0.07 & 0.04 & \bftab0.01 & 0.06 & 0.05 & 0.10 \\
                             & \cellcolor{light-gray}ResNet-50+Alignment+DM    &\cellcolor{light-gray}\bftab{0.08} &
                                                            \cellcolor{light-gray}\bftab{0.41} & 
                                                            \cellcolor{light-gray}\bftab{0.11} &
                                                            \cellcolor{light-gray}0.11 & 
                                                            \cellcolor{light-gray}0.10 &
                                                            \cellcolor{light-gray}\bftab{0.01} & 
                                                            \cellcolor{light-gray}\bftab{0.08} &
                                                            \cellcolor{light-gray}\bftab{0.11} &
                                                            \cellcolor{light-gray}\bftab{0.13}             
                                                            \\ \Xhline{2\arrayrulewidth}
\end{tabular}
}
\end{table}
We compared the results of our method with those of previous models~\cite{wang2017chestx, cai2018iterative,liu2019align} on eight different thoracic diseases. The quantitative evaluation results of the localization performances are reported in Table~\ref{tab1}. Our proposed model achieved improvements of 9\% and 3\% compared with the previous state-of-the-art localization performances at IoU of 0.3 and 0.5, respectively. Additionally, we compared three types of models to analyze the effects of the alignment module and disease masks on the deep network's localization performance. The base model is the ResNet-50 network that we used as a feature extractor. The other models are a ResNet-50 trained with aligned X-ray images and our proposed model, which implements spatial attention using a disease mask with aligned X-ray images. In Table \ref{tab1}, our model showed significantly improved localization performance compaired with the ResNet-50 and ResNet-50+Alignment models for large diseases such as `cardiomegaly' or `pneumothorax.' However, for small diseases such as `nodule', our model showed relatively little improvement. This result indicates that even though the disease mask serves to focus the network's attention to the disease's locations, the effect is trivial on small diseases because it guides the attention to a relatively large area compared to the size of the disease area. On the other hand, for a large diseases, the disease mask approach achieves excellent localization performance because it focuses the overall attention on the disease occurrence area. 
\begin{sidewaysfigure}[p]
\centering
\includegraphics[width=1.0\columnwidth]{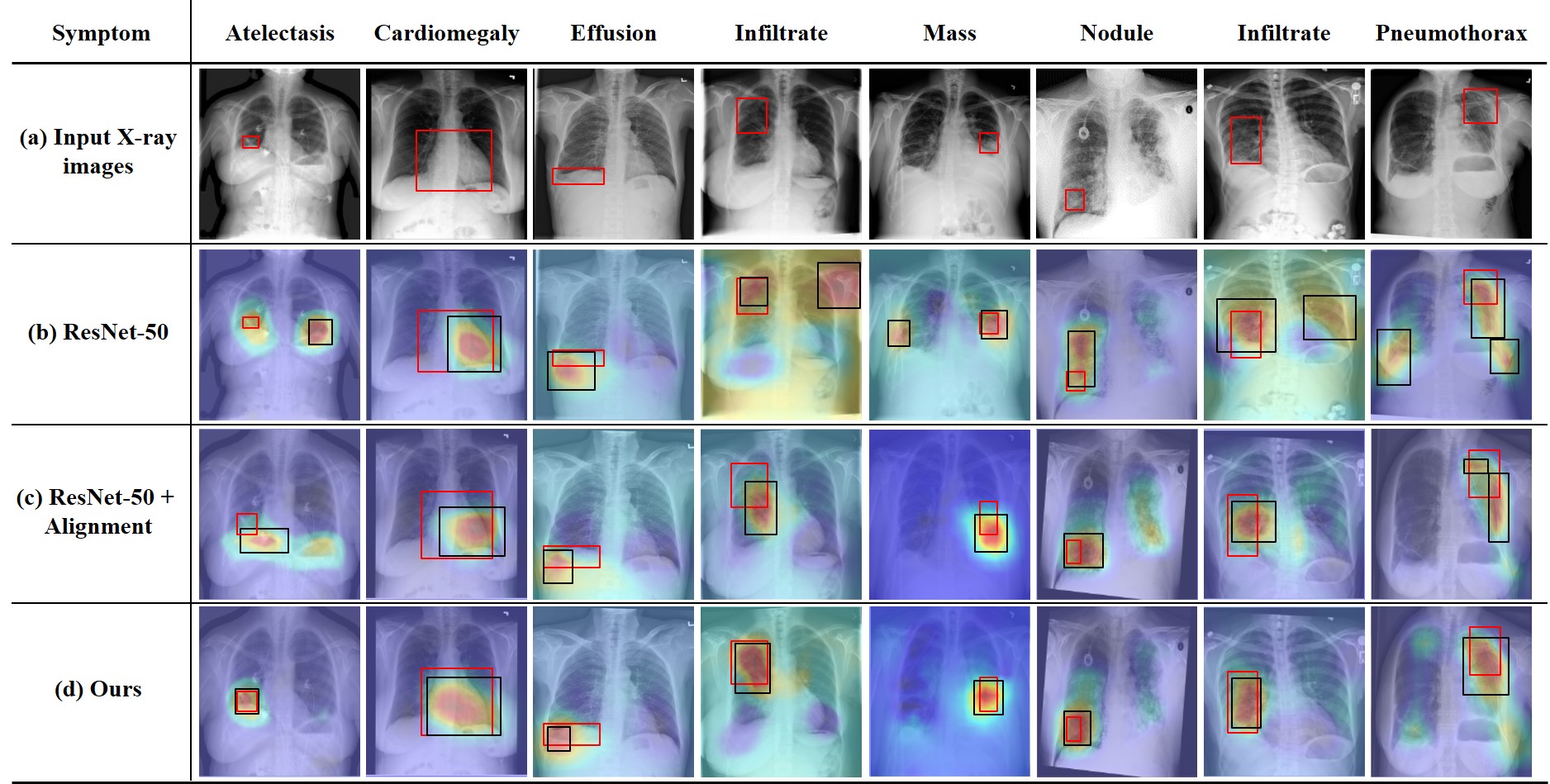}
\vspace*{-10mm}
\caption{Qualitative evaluation of thoracic disease localization for three tested models. We compare our proposed method with ResNet-50 and ResNet-50+alignment models on eight types of disease classes. The columns represent the symptoms of thoracic diseases. The red and black boxes indicate the ground truth and the predicted box from the classifier, respectively.
}\label{fig7}
\end{sidewaysfigure}

Fig. \ref{fig7} shows qualitative evaluation results for three types of models on the eight types of thoracic diseases. Row (a) represents the preprocessed original X-ray image. Rows (b-d) show a comparison of the localization performance by each model. Our proposed method shows improved disease localization results compared with the baselines. The cardiomegaly prediction box of (b-c) is skewed to the right compared to the ground truth box. However, the prediction box (d) shows that the mask guides the class activation map to the exact heart disease area. As another example, the infiltrate's prediction box of (b) shows localizing the wrong area outside the lungs. However, the infiltrate occurs in the lungs of the chest. Our masks provided this prior information to the network, and the (d) row shows more accurate localization results.

\begin{figure}[t!]
\centering
  \includegraphics[width=0.8\columnwidth]{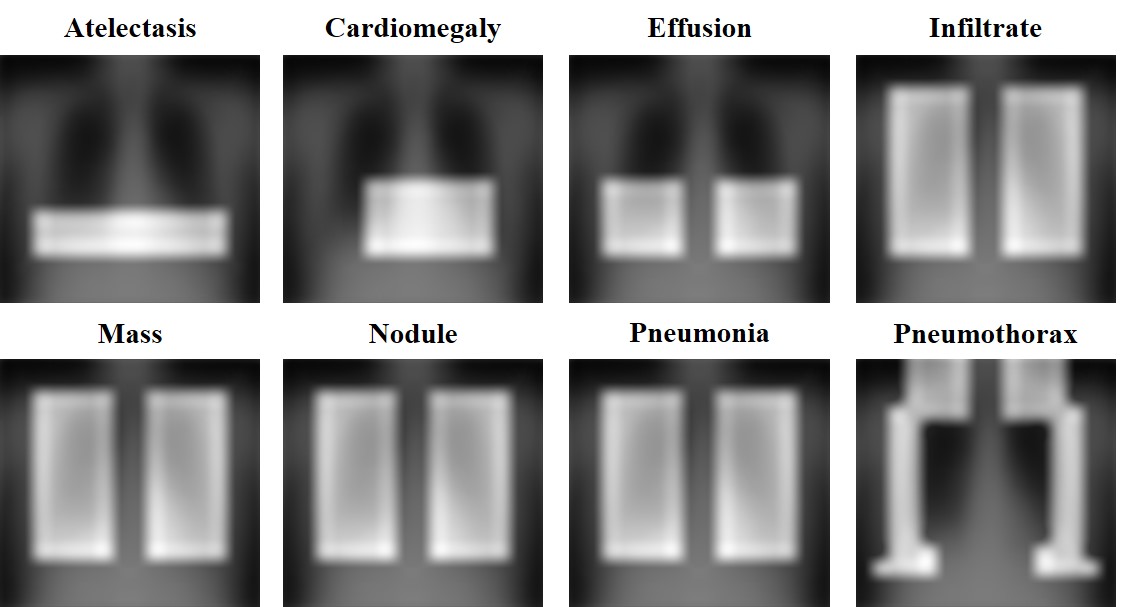}
   \caption{Pseudo-disease masks based on external knowledge. The pseudo-disease masks are marked using an anchor X-ray image. In each pseudo-disease mask, the disease area is filled with `1's, and the nondisease area is filled with `0's.
}\label{fig8}
\end{figure}

\begin{table}[t!]
\caption{Localization performances by networks trained using PDM and DM. We used a ResNet-50 pretrained on aligned X-ray images as a base model. Our disease masks describe the representative area for each disease in more detail than do the pseudo-disease masks.}
\label{tab2}
\centering
\resizebox{\textwidth}{!}{
\begin{tabular}{l|l|llllllll|l}
\Xhline{2\arrayrulewidth}
\multicolumn{1}{c|}{T(IoU)} & \multicolumn{1}{c|}{Model} & \multicolumn{1}{c}{Ate.} & \multicolumn{1}{c}{Car.} & \multicolumn{1}{c}{Eff.} & \multicolumn{1}{c}{Inf.} & \multicolumn{1}{c}{Mas.} & \multicolumn{1}{c}{Nod.} & \multicolumn{1}{c}{Pn1.} & \multicolumn{1}{c|}{Pn2.} & \multicolumn{1}{c}{Mean} \\ \Xhline{2\arrayrulewidth}
\multicolumn{1}{c|}{0.3} & Base model+PDM  & 0.34 & 0.90 & \bftab{0.44} & 0.45 & 0.43 & 0.12 & 0.31 & \bftab{0.45} & 0.43 \\ \cline{2-11} 
                             & \cellcolor{light-gray}Base model+DM    &\cellcolor{light-gray}\bftab{0.40} & 
                                                            \cellcolor{light-gray}\bftab{0.97} &
                                                            \cellcolor{light-gray}0.40 &
                                                            \cellcolor{light-gray}\bftab{0.58} &
                                                            \cellcolor{light-gray}\bftab{0.50} &
                                                            \cellcolor{light-gray}\bftab{0.14} & 
                                                            \cellcolor{light-gray}\bftab{0.40} & 
                                                            \cellcolor{light-gray}0.38 & 
                                                            \cellcolor{light-gray}\bftab{0.47}            
                            \\ \hline \hline
\multicolumn{1}{c|}{0.5}    & Base model+PDM   & 0.16 & 0.75 & \bftab{0.22} & 0.20 & 0.30 & 0.03 & 0.09 & \bftab{0.20} & 0.24 \\ \cline{2-11} 
                             & \cellcolor{light-gray}Base model+DM    &\cellcolor{light-gray}\bftab{0.18} &
                                                            \cellcolor{light-gray}\bftab{0.80} &
                                                            \cellcolor{light-gray}0.21 &
                                                            \cellcolor{light-gray}\bftab{0.25} &
                                                            \cellcolor{light-gray}\bftab{0.35} & 
                                                            \cellcolor{light-gray}\bftab{0.05} & 
                                                            \cellcolor{light-gray}\bftab{0.15}& 
                                                            \cellcolor{light-gray}0.16 &
                                                            \cellcolor{light-gray}\bftab{0.27}
                            \\ \hline \hline
\multicolumn{1}{c|}{0.7}     & Base model+PDM  & 0.07 & 0.40 & \bftab{0.12} & 0.08 & 0.05 & \bftab0.01 & 0.05 & \bftab{0.12} & 0.11 \\ \cline{2-11} 
                             & \cellcolor{light-gray}Base model+DM    &\cellcolor{light-gray}\bftab{0.08} &
                                                            \cellcolor{light-gray}\bftab{0.41} & 
                                                            \cellcolor{light-gray}0.11 &
                                                            \cellcolor{light-gray}\bftab{0.11} & 
                                                            \cellcolor{light-gray}\bftab{0.10} &
                                                            \cellcolor{light-gray}\bftab{0.01} & 
                                                            \cellcolor{light-gray}\bftab{0.08} &
                                                            \cellcolor{light-gray}0.11 &
                                                            \cellcolor{light-gray}\bftab{0.13}             
                                                            \\ \Xhline{2\arrayrulewidth}
\end{tabular}
}
\end{table}

\begin{figure}[t!]
\centering
  \includegraphics[width=0.8\columnwidth]{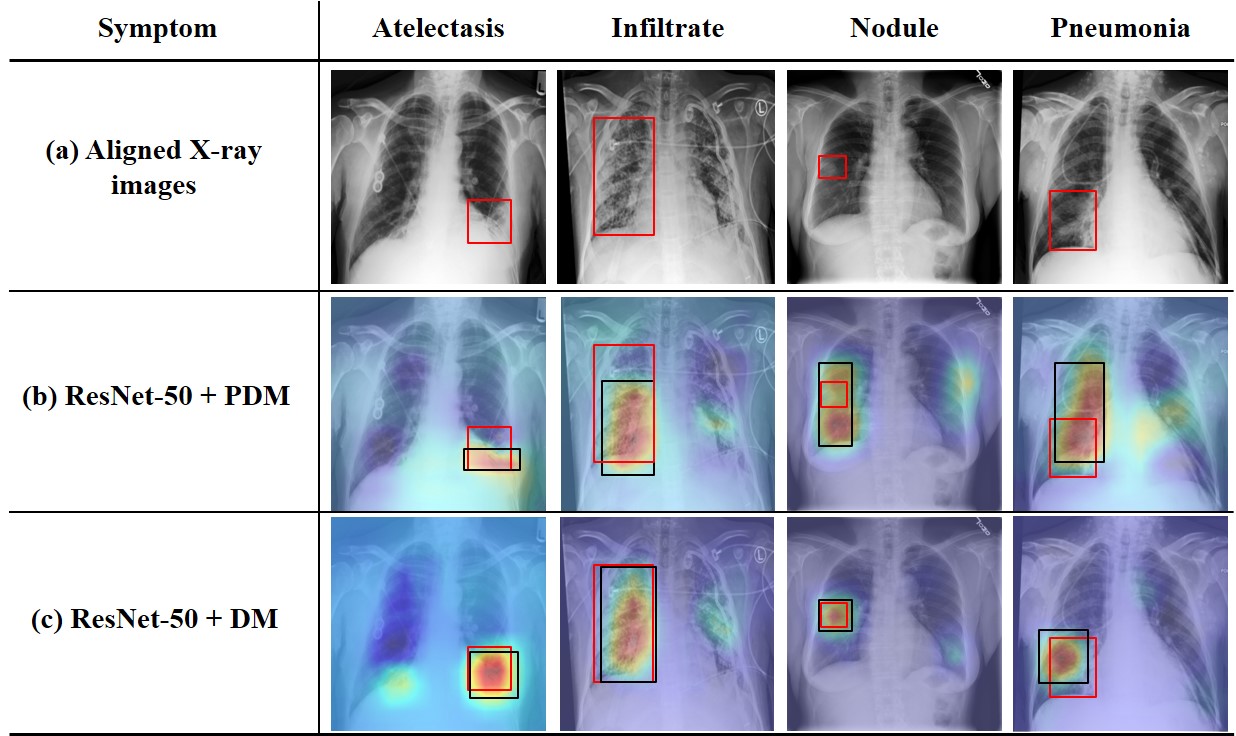}
   \caption{Qualitative evaluation of thoracic disease localization for networks trained with PDMs and DMs. The network trained with DM localize more accurate areas than does the network trained with PDM. The red and black boxes indicate the ground truth and the predicted box from the classifier, respectively.
}\label{fig9}
\end{figure}

\subsection{Pseudo-disease masks based on external knowledge}
In this section, we conduct additional comparative experiments to demonstrate the effectiveness of the disease mask approach. We proposed a spatial attention method using the disease masks that included the disease area information. However, disease area information can easily be found via the Internet or in medical books. Thus, we manually generated pseudo-disease masks (PDMs) based on external knowledge from Wikipedia and conducted comparative experiments to demonstrate that our method generates more specific disease areas than pseudo ones. Fig.~\ref{fig8} shows the pseudo-disease masks marked above the anchor images. The pseudo disease masks are filled with `1' in the disease area and `0' in the nondisease area. `Infiltrate', `mass', `nodule', and `pneumonia' can occur everywhere in the lung area; therefore, we marked all the lung areas for those four diseases. We trained the base model using these pseudo-masks and compared the localization performance with our method. In Table~\ref{tab2}, the base model trained using the proposed disease mask showed 4\%, 3\%, and 2\% higher localization performance at IoUs of 0.3, 0.5, and 0.7. The results indicate that the generated masks more effectively guide the location of the disease than do the pseudo-disease masks manually generated based on external knowledge. The qualitative disease localization evaluation scores of the networks using the pseudo disease and proposed disease masks are reported in Fig.~\ref{fig9}. We can observe that the model using pseudo-disease masks can find the surrounding area of diseases but unlike the proposed method, it has difficulty finding precise bounding boxes. 

\begin{table*}[t!]
\caption{A comparison of the classification performance on 14 types of diseases using the AUC metric. Asterisks (*) denote methods that use additional bounding box information during training}
\label{tab3}
\centering
\resizebox{\textwidth}{!}{
\begin{tabular}{cccccc}
\Xhline{2\arrayrulewidth}
\multicolumn{1}{c||}{Pathology}& \multicolumn{1}{c}{X. Wang \textit{et al}.~\cite{wang2017chestx}} & \multicolumn{1}{c}{J. Cai \textit{et al}.~\cite{cai2018iterative}} & \multicolumn{1}{c}{*L. Zhe \textit{et al}.~\cite{li2018thoracic}} & \multicolumn{1}{c}{*J. Liu \textit{et al}.~\cite{liu2019align}} & \multicolumn{1}{c}{Ours} \\ \Xhline{2\arrayrulewidth}
\multicolumn{1}{c||}{Atelectasis}        & 0.70           & 0.75          & \bftab 0.80          & 0.79          & 0.79 \\ \hline
\multicolumn{1}{c||}{Cardiomegaly}       & 0.81           & 0.86          & 0.87          & 0.87          & \bftab 0.91 \\ \hline
\multicolumn{1}{c||}{Effusion}           & 0.76          & 0.81          & 0.87          & \bftab 0.88          & 0.84 \\ \hline
\multicolumn{1}{c||}{Infiltration}       & 0.66           & 0.67          & 0.70          & 0.69          & \bftab 0.71 \\ \hline
\multicolumn{1}{c||}{Mass}               & 0.69           & 0.80          & 0.83          & 0.81          & \bftab 0.85 \\ \hline
\multicolumn{1}{c||}{Nodule}             & 0.67           & 0.76          & 0.75          & 0.73          & \bftab 0.81 \\ \hline
\multicolumn{1}{c||}{Pneumonia}          & 0.66           & 0.70          & 0.67          & \bftab 0.75          & 0.73 \\ \hline
\multicolumn{1}{c||}{Pneumothorax}       & 0.80           & 0.85          & 0.87          & \bftab 0.89          & \bftab 0.89 \\ \hline
\multicolumn{1}{c||}{Consolidation}      & 0.70           & 0.69          & \bftab 0.80          & 0.79          & 0.74 \\ \hline
\multicolumn{1}{c||}{Edema}              & 0.81           & 0.83          & 0.88          & \bftab 0.91          & 0.86 \\ \hline
\multicolumn{1}{c||}{Emphysema}          & 0.83           & 0.91          & 0.91          & 0.93          & \bftab 0.95 \\ \hline
\multicolumn{1}{c||}{Fibrosis}           & 0.79           & 0.80          & 0.78          & 0.80          & \bftab 0.84 \\ \hline
\multicolumn{1}{c||}{Pleural Thickening} & 0.68           & 0.75          & 0.79          & \bftab 0.80          & \bftab 0.80 \\ \hline
\multicolumn{1}{c||}{Hernia}             & 0.87           & 0.53          & 0.77          & \bftab 0.92          & 0.88 \\ \Xhline{2\arrayrulewidth}
\rowcolor{light-gray}\multicolumn{1}{c||}{Mean}               & 0.74           & 0.77          & 0.81          & \bftab 0.83          & \bftab 0.83 \\ \Xhline{2\arrayrulewidth}
\end{tabular}
}
\end{table*}
\begin{table*}[t!]
\caption{The results of an ablation study showing the effect of each proposed module on the classification performance. We apply the models to eight types of diseases that are the same ones evaluated in the localization performance experiments.}
\label{tab4}
\centering
\resizebox{\textwidth}{!}{
\begin{tabular}{l|llllllll|l}
\Xhline{2\arrayrulewidth}
\multicolumn{1}{c|}{Model}              & \multicolumn{1}{l}{Ate.} & \multicolumn{1}{l}{Car.} & \multicolumn{1}{l}{Eff.} & \multicolumn{1}{l}{Inf.} & \multicolumn{1}{l}{Mas.} & \multicolumn{1}{l}{Nod.} & \multicolumn{1}{l}{Pn1.} & Pn2. & Mean  \\ \Xhline{2\arrayrulewidth}
ResNet-50 & 0.75 & 0.90 & 0.80 & 0.70 & 0.81 & 0.78 & 0.71 & 0.86 & 0.79 \\ \hline
ResNet-50+Alignment & 0.77 & 0.90 & 0.83 & 0.70 & 0.82 & 0.79 & 0.72 & 0.88 & 0.80 \\ \hline 
\rowcolor{light-gray} ResNet-50+Alignment+DM               & \bftab 0.79                      & \bftab 0.91                      & \bftab 0.84                      & \bftab 0.71                      & \bftab 0.85                      & \bftab 0.81                      & \bftab 0.73                      & \bftab 0.89 & \bftab 0.82 \\ \Xhline{2\arrayrulewidth}
\end{tabular}
}
\end{table*}

\subsection{Evaluations of thoracic diseases classification}
Because our method aims to improve the localization performance by applying spatial attention to a classifier, we further provide the classification performance using the Area Under the Curve (AUC) metric for 14 types of diseases. As shown in Table~\ref{tab3}, the proposed method outperforms the state-of-the-art methods~\cite{wang2017chestx, cai2018iterative, li2018thoracic}. Although the method of J. Liu \textit{et al}. \cite{liu2019align} showed an AUC performance equivalent to that of our method, they exploit additional bounding box information during training unlike ours. Furthermore, the localization performances in Table \ref{tab1} demonstrate that using spatial attention in conjunction with disease masks is helpful for explicitly focusing a classifier on the disease areas even with the comparable capacity of classifiers. 

In Table~\ref{tab4}, we provide ablation studies that show how the disease masks and the alignment module affect the AUC performance. We can observe that the alignment module removes noise from the X-ray images, while the attention method is crucial for improving the classification performance.


\section{Conclusion}
In this study, we proposed a spatial attention method to train a deep network that considers the distributions of disease areas. Disease masks were generated by preprocessing the feature maps of a network pretrained on the training data. Furthermore, we introduced a unified framework to incorporate the disease masks, an alignment module and weighted cross-entropy loss to enhance the quality of X-ray images and reduce skewed predictions. Finally, we showed that the retrained feature extraction method with the unified framework improved the localization of thoracic diseases on the ChestX-ray14 dataset. Moreover, we verified the quality of the generated masks by comparing their localization performances with those of pseudo-disease masks obtained by external knowledge. Finally, the proposed method achieved state-of-the-art classification performance without additional instance-level annotations, which indicates the effectiveness of spatial attention with the disease masks approach.

\section*{Acknowledgment}
This work was supported by Institute for Information \& communications Technology Planning \& Evaluation(IITP) grant funded by the Korea government(MSIT) (No. 2017-0-01779, A machine learning and statistical inference framework for explainable artificial intelligence and No. 2019-0-00079, Artificial Intelligence Graduate School Program, Korea University).

\bibliography{mybibfile}

\end{document}